\documentclass[runningheads]{llncs}

\usepackage{multirow}
\usepackage{eccv}

\usepackage{eccvabbrv}

\usepackage{graphicx}
\usepackage{booktabs}
\usepackage{algorithm}
\usepackage[table]{xcolor}
\usepackage{algpseudocode}
\usepackage[accsupp]{axessibility}  

\usepackage{hyperref}

\usepackage{orcidlink}

\begin{document}

\title{EmbodiedHead: Real-Time Listening and Speaking Avatar for Conversational Agents} 

\titlerunning{EmbodiedHead}


\newcommand{\equalcontribsymbol}{\ensuremath{*}}
\newcommand{\correspondingsymbol}{\ensuremath{\dagger}}
\newcommand{\authormarker}[1]{\,\textsuperscript{\normalfont#1}}

\author{
Yu Zhang\inst{1}\orcidlink{0009-0000-1833-3012}%
\authormarker{\equalcontribsymbol}
\and
Kaiyuan Shen\inst{1}\orcidlink{0009-0006-6747-1255}%
\authormarker{\equalcontribsymbol}
\and
Yang Li\inst{1,2}\orcidlink{0000-0001-9427-7665}%
\authormarker{\correspondingsymbol}
}

\authorrunning{Zhang et al.}

\institute{
School of Computer Science and Technology,
East China Normal University, Shanghai, China\\
\email{\{51275900029,51285900014\}@stu.ecnu.edu.cn,
yli@cs.ecnu.edu.cn}
\and
Garabido Intelligence, Shanghai, China
}

\maketitle

\begingroup
\renewcommand{\thefootnote}{\equalcontribsymbol}
\footnotetext[1]{Equal contribution.}

\renewcommand{\thefootnote}{\correspondingsymbol}
\footnotetext[2]{Corresponding author.}
\endgroup

\begin{center}
\small
\href{https://vpx-ecnu.github.io/EmbodiedHead-Website/}{\texttt{\textcolor{eccvblue}{https://vpx-ecnu.github.io/EmbodiedHead-Website/}}}
\end{center}

\begin{abstract}
We present EmbodiedHead, a speech-driven talking head framework that equips LLMs with real-time visual avatars for conversation. A practical embodied avatar must achieve real-time generation, unified listening-speaking behavior, and high rendered visual quality simultaneously. Our framework couples the first Rectified-Flow Diffusion Transformer (DiT) for this task with a differentiable renderer, enabling diverse, high-fidelity generation in as few as four sampling steps. Prior listening-speaking methods rely on dual-stream audio, introducing an interlocutor look-ahead dependency incompatible with causal user--LLM interaction. We instead adopt a single-stream interface with explicit per-frame listening-speaking state conditioning and a Streaming Audio Scheduler, suppressing spurious mouth motion during listening while enabling seamless turn-taking. A two-stage training scheme of coefficient-space pretraining and joint image-domain refinement further closes the gap between motion-level supervision and rendered quality. Extensive experiments demonstrate state-of-the-art visual quality and motion fidelity in both speaking and listening scenarios.

\end{abstract}

\section{Introduction}
\label{sec:intro}

Large Language Models (LLMs) are now widely used for daily chat and assistance, but most systems still interact through plain text or voice. Without a visible face, users miss social cues such as eye contact, facial expression, and head motion that convey attention, emotion, and turn-taking. Embodied Social Presence (ESP) Theory argues that an embodied representation (\eg, an avatar) becomes the focal point through which people experience co-presence and interpret social interaction; richer embodied cues strengthen perceived social presence and engagement in mediated communication~\cite{5428431}.

In this paper we study \emph{Head-Embodied LLMs}: equipping an LLM with a head-only visual avatar that listens and speaks in real time during conversation. A practical head-embodied LLM avatar must meet three goals at once: (i) real-time generation to support natural turn-taking; (ii) unified listening-speaking behavior to provide role-aware nonverbal signaling throughout the full interaction rather than only during speech; and (iii) high rendered visual quality to ensure perceptual plausibility and prevent visual artifacts from undermining credibility and social presence.
Compared to 2D-based methods \cite{jiangLoopyTamingAudioDriven2024,cuiHallo3HighlyDynamic2025}, 3D-based methods~\cite{Fan_2022_CVPR,Xing_2023_CVPR,sunDiffPoseTalkSpeechDrivenStylistic2024,chuARTalkSpeechDriven3D2025} generally enable lower inference cost, making them a promising route toward Head-Embodied LLMs.

Most current listening-speaking integrated methods \cite{pengDualTalkDualSpeakerInteraction2025,chenSeamlessInteractionCausal2025,chuUniLSEndtoEndAudioDriven2025,zhuMANGONaturalMultispeaker2026} employ a dual-stream audio architecture, whose datasets suffer from limited diversity in languages and scenarios.
More importantly, dual-stream conditioning introduces an interlocutor look-ahead dependency: generating the current avatar response requires the interlocutor's reactive audio or visual feedback within the same window, which is not available at inference time in user--LLM interactions. Consequently, the model cannot reliably leverage cross-stream cues in a causal setting, and the dual-stream design may collapse into an effectively single-audio, alternately driven paradigm.
Beyond the interaction architecture, most 3D talking-head methods treat mesh generation and image rendering as two separate steps. Motion is trained and evaluated in coefficient space, so the reported speed and quality numbers do not reflect the rendered visual quality that a head-embodied LLM avatar actually delivers to the user.

\begin{figure}[tbp]
\includegraphics[width=12cm]{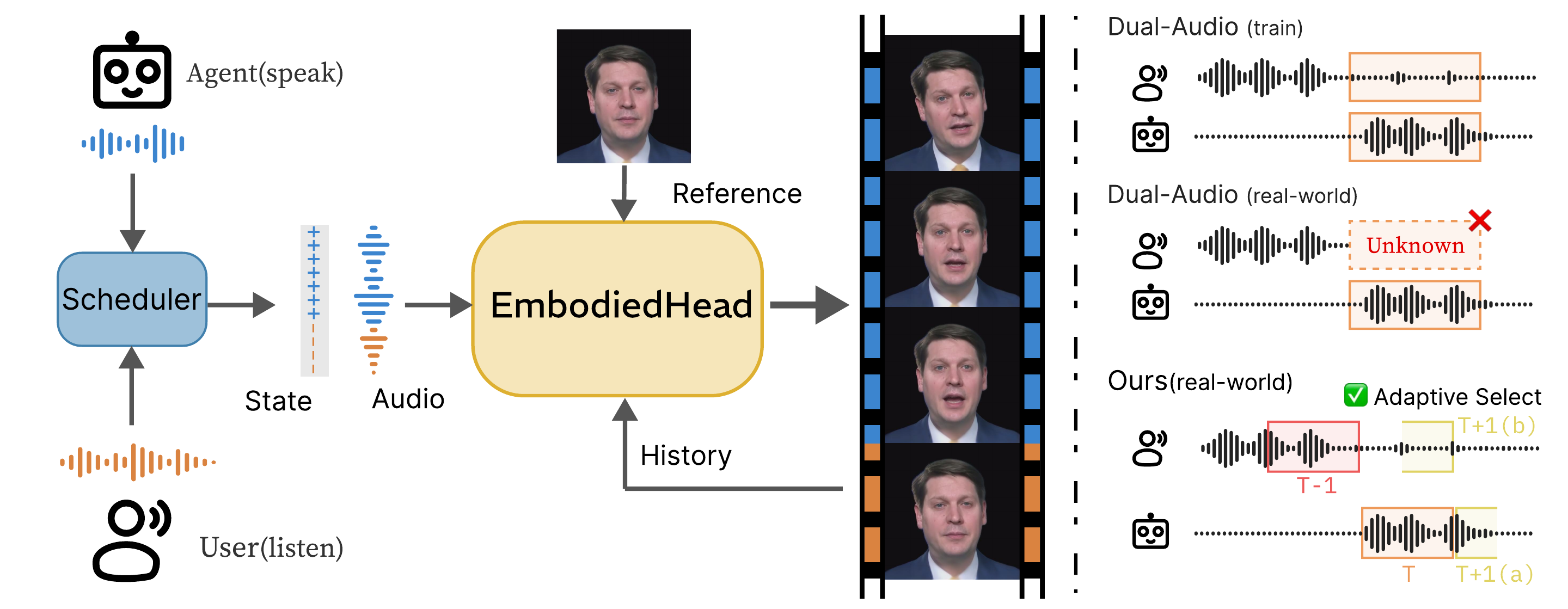}  
\caption{We present EmbodiedHead, which generates a real-time head-embodied avatar for LLMs. Unlike dual-audio methods, it uses a single audio stream with explicit listening-speaking state conditioning to achieve unified conversational behavior.}
\label{fig:example}
\end{figure}

To address these issues, we propose \textbf{EmbodiedHead}, an end-to-end speech-driven talking-head framework designed for head-embodied LLMs (\cref{fig:example}). Our pipeline couples the first Rectified-Flow~\cite{liu2022flowstraightfastlearning} Diffusion Transformer (DiT)~\cite{Peebles_2023_ICCV,chen2023pixartalphafasttrainingdiffusion} for this task with a differentiable renderer, retaining diffusion-model diversity while enabling high-fidelity generation in as few as four sampling steps.
To remove the interlocutor look-ahead dependency, we adopt a single-stream audio interface and inject an explicit per-frame \emph{listening-speaking state} (LS-state) into both the model input and the audio cross-attention via FiLM modulation~\cite{perez2018film}, giving the model a clear mode signal at every frame. A \emph{Streaming Audio Scheduler} merges microphone and LLM audio into a unified window and emits aligned LS-states, enabling rapid turn-taking without future information.
To close the gap between coefficient-space supervision and rendered quality, we adopt a two-stage training scheme: the DiT is first pretrained with flow matching in coefficient space for stable dynamics, then jointly fine-tuned with the renderer using image-domain losses. The straight-path property of Rectified Flow supports reliable one-step endpoint estimation, making this joint optimization both well-grounded and efficient.

Overall, the contributions of this paper can be summarized as follows:
\begin{itemize}
\item[$\bullet$] We propose an end-to-end framework coupling a Rectified-Flow DiT with a differentiable renderer and multi-level conditioning, achieving real-time, diverse talking-head generation in few sampling steps.
\item[$\bullet$] We enable unified listening-speaking behavior via explicit LS-state conditioning and a Streaming Audio Scheduler, suppressing listening-phase mouth hallucinations and supporting rapid turn-taking.
\item[$\bullet$] A two-stage scheme jointly optimizes the DiT and renderer with image-domain supervision, closing the gap between coefficient-space training and rendered quality. Extensive experiments demonstrate state-of-the-art performance in both speaking and listening scenarios.
\end{itemize}

\section{Related Work}
\label{sec:related}

\subsection{Speech-driven 3D Talking Head Generation}

Most work in speech-driven 3D talking-head generation follows a two-stage pipeline: a motion model generates 3D motion from audio, which is then rendered by a separate module~\cite{Cudeiro_2019_CVPR,Fan_2022_CVPR,Xing_2023_CVPR,sunDiffPoseTalkSpeechDrivenStylistic2024}.
VOCA~\cite{Cudeiro_2019_CVPR} established identity-decoupled regression from speech. FaceFormer~\cite{Fan_2022_CVPR} and CodeTalker~\cite{Xing_2023_CVPR} introduced Transformer modeling and discrete motion priors to improve lip sync and motion diversity. DiffPoseTalk~\cite{sunDiffPoseTalkSpeechDrivenStylistic2024} applied diffusion for style-conditioned generation, while ARTalk~\cite{chuARTalkSpeechDriven3D2025} and GLDiTalker~\cite{lin2025glditalkerspeechdriven3dfacial} target real-time output via autoregressive codebooks and latent diffusion. Despite these advances, training and evaluation stay in coefficient space: the reported accuracy and speed reflect mesh-level performance, not the rendered visual quality that users actually perceive.

A parallel line of work bypasses parametric models and drives a per-identity neural scene directly with audio, using either NeRF~\cite{mildenhall2020nerfrepresentingscenesneural,Guo_2021_ICCV,liEfficientRegionAwareNeRF2023} or 3D Gaussian Splatting~\cite{kerbl3Dgaussians,aneja2024gaussianspeechaudiodrivengaussianavatars}. These methods can capture fine appearance details, but the reconstructed scene is largely static: motion is limited to the lip region while eye movement, head pose, and lighting variation are poorly handled, yielding a rigid and visually flat result. Per-identity scene reconstruction also adds a costly setup stage, reducing suitability for general head-embodied LLM deployment.
Some 2D methods learn a direct mapping from audio to images and achieve high rendered quality~\cite{jiangLoopyTamingAudioDriven2024,cuiHallo3HighlyDynamic2025}, but their generation speed is too slow for real-time head-embodied LLM interaction.

\subsection{Listening-Speaking Integrated 3D Avatar Generation}

Unified listening-speaking generation has evolved from single-listener modeling toward multi-turn streaming frameworks~\cite{Ng_2022_CVPR,pengDualTalkDualSpeakerInteraction2025,chenSeamlessInteractionCausal2025,chuUniLSEndtoEndAudioDriven2025,zhuMANGONaturalMultispeaker2026,tran2024,siniukov2025ditailistenercontrollablehighfidelity}. Learning to Listen~\cite{Ng_2022_CVPR} modeled stochastic 3D listener responses but did not address multi-turn dialogue. DualTalk~\cite{pengDualTalkDualSpeakerInteraction2025} extended this to multi-round conversations under a dual-stream setting, but processes complete audio sequences offline and generates outputs in a single pass, making streaming deployment infeasible. TIMAR~\cite{chenSeamlessInteractionCausal2025} introduced turn-level causal modeling to address latency, yet relies on the interlocutor's visual stream as input, which adds extra acquisition cost in real-time scenarios. INFP~\cite{Zhu_2025_CVPR} and UniLS~\cite{chuUniLSEndtoEndAudioDriven2025} further improved naturalness and continuity under dual-stream audio settings.
Despite this progress, all dual-stream methods share a common structural limitation: generating the current window requires the interlocutor's reactive audio for the same span, which is unavailable in real user--LLM chat. This interlocutor look-ahead dependency makes dual-stream conditioning incompatible with causal streaming inference, and in practice these methods tend to degrade into alternated single-audio driving with unreliable mode inference during listening phases.
MANGO~\cite{zhuMANGONaturalMultispeaker2026} recognized the value of image-domain supervision and introduced 2D-lifted training alongside dual-stream audio. However, MANGO is built on DDPM~\cite{ho2020denoising}, which requires many sampling steps to produce reliable outputs. Single-step estimates from DDPM carry large approximation errors, so image-domain losses cannot be grounded in outputs that match actual inference behavior, limiting the practical effect of the image supervision.

\section{Method}
\label{sec:method}

\begin{figure}[tbp]
\includegraphics[width=\linewidth]{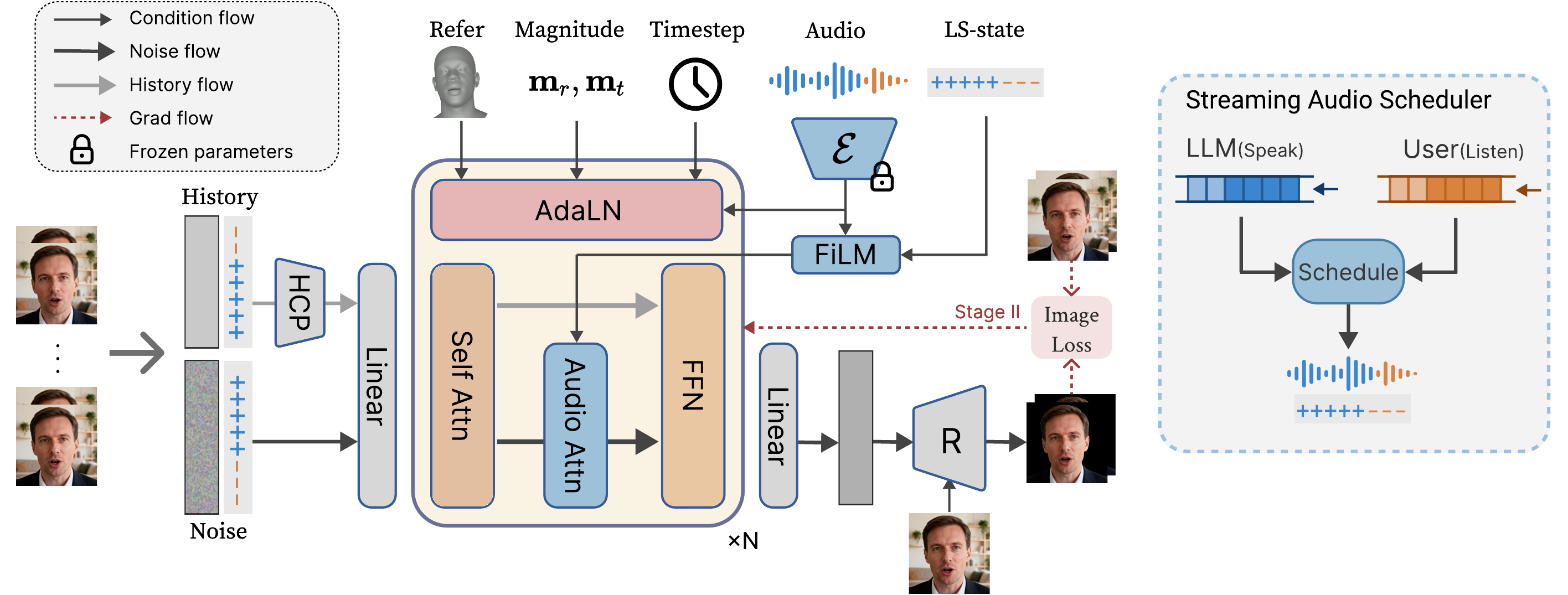}  
\caption{EmbodiedHead employs a Rectified-Flow DiT to generate speech-driven talking-head animation in few steps. It conditions on reference, timestep, motion magnitude, and LS-state. A streaming scheduler merges user--LLM audio, enabling unified listening-speaking.}
\label{fig:method_pic}
\end{figure}

\cref{fig:method_pic} provides an overview of EmbodiedHead. After formalizing the task and introducing Rectified Flow (\cref{sec:prelim}), we detail the injection of multiple conditions into DiT blocks (\cref{sec:mc_dit}). To unify speaking and listening behaviors in conversational streams, we then propose a listening-speaking state (\cref{sec:ls_control}), culminating in a two-stage training scheme for end-to-end image-domain refinement (\cref{sec:two_stage}).

\subsection{Preliminaries}
\label{sec:prelim}

Given an input audio segment $\mathbf{a}$, we aim to generate a temporally coherent 3D motion sequence $\mathbf{x}_1\in\mathbb{R}^{T\times D}$ for a streaming window of length $T$ frames. We use $\mathbf{c}$ to denote the full set of conditioning variables.

We use FLAME~\cite{FLAME:SiggraphAsia2017} to represent facial geometry, parameterizing each frame by an expression vector $\mathbf{e}^{\tau} \in \mathbb{R}^{100}$ and a pose vector $\mathbf{p}^{\tau}$ for $\tau\in\{1,\ldots,T\}$, with a shared identity shape $\mathbf{s}\in\mathbb{R}^{300}$. The per-frame motion vector is defined as
\begin{equation}
\mathbf{x}_1^{\tau} = [\mathbf{e}^{\tau},\;\mathbf{p}^{\tau}],\qquad D=100+\dim(\mathbf{p}^{\tau}).
\end{equation}

We adopt Rectified Flow~\cite{liu2022flowstraightfastlearning} to retain the diversity of diffusion models while reducing sampling steps. Let $\mathbf{x}_1$ be a data sample and $\mathbf{x}_0\sim\mathcal{N}(\mathbf{0},\mathbf{I})$ be Gaussian noise. Rectified Flow defines a straight interpolation path
\begin{equation}
\mathbf{x}_t = \mathbf{x}_0 + t\,(\mathbf{x}_1-\mathbf{x}_0),\qquad t\sim\mathcal{U}[0,1].
\label{eq:rf_path}
\end{equation}
Along this path, the target velocity is a constant vector $\mathbf{u}_t = \frac{d\mathbf{x}_t}{dt} = \mathbf{x}_1-\mathbf{x}_0$. We train a conditional velocity field $\mathbf{v}_\theta(\mathbf{x}_t,t,\mathbf{c})$ by minimizing the flow-matching objective~\cite{lipman2023flowmatchinggenerativemodeling}
\begin{equation}
\mathcal{L}_{\mathrm{FM}} = \mathbb{E}\Big[\big\|\mathbf{v}_\theta(\mathbf{x}_t,t,\mathbf{c})-(\mathbf{x}_1-\mathbf{x}_0)\big\|_2^2\Big].
\label{eq:fm_loss}
\end{equation}

\subsection{Multi-Condition Rectified-Flow DiT}
\label{sec:mc_dit}

We parameterize $\mathbf{v}_\theta$ with a DiT to model the conditional velocity field. The condition is instantiated as
$\mathbf{c}=(\mathbf{a},\mathbf{s},\mathbf{x}^{\mathrm{ref}},\mathbf{x}^{\mathrm{hist}},\mathbf{q},\mathbf{m})$,
where $\mathbf{x}^{\mathrm{ref}}$ and $\mathbf{s}$ represent the reference frame, $\mathbf{x}^{\mathrm{hist}}$ denotes history FLAME parameters, $\mathbf{q}$ is the per-frame listening-speaking state (\cref{sec:ls_control}) and $\mathbf{m}$ controls normalized rotation/translation magnitudes. To inject these conditions in a stable and controllable manner, we adopt three complementary pathways.

\subsubsection{Model input.}
We build a single sequence of motion tokens by concatenating a history FLAME $\mathbf{x}^{\mathrm{hist}}\in\mathbb{R}^{L\times D}$ and the current noisy window $\mathbf{x}_{t}\in\mathbb{R}^{T\times D}$, and explicitly append the listening-speaking state $\mathbf{q}=[\mathbf{q}^{\mathrm{hist}},\mathbf{q}^{\mathrm{cur}}]\in\{-1,+1\}^{L+T}$ as an extra channel to every frame feature. Given the flow time $t$, we form augmented frame features by channel-wise concatenation
\begin{equation}
\tilde{\mathbf{x}}^{\mathrm{hist}}=[\mathbf{x}^{\mathrm{hist}},\mathbf{q}^{\mathrm{hist}}],\qquad\tilde{\mathbf{x}}_{t}=[\mathbf{x}_{t},\mathbf{q}^{\mathrm{cur}}].
\end{equation}
Inspired by~\cite{zhang2025framepack}, we propose History Context Packing (HCP) to efficiently encode long-range context: $\tilde{\mathbf{x}}^{\mathrm{hist}}$ is partitioned into $G$ groups of exponentially growing temporal spans, and each group is compressed into a fixed number of tokens via a learnable linear projection, yielding $\tilde{\mathbf{h}}\in\mathbb{R}^{H\times(D+1)}$. Finally, we concatenate these packed history tokens with the current tokens and project them to the transformer dimension: $\mathbf{Z}_0=\mathrm{Linear}([\tilde{\mathbf{h}},\tilde{\mathbf{x}}_{t}])\in\mathbb{R}^{(H+T)\times d}$.

\subsubsection{Frame-level audio attention.}
We use mHuBERT~\cite{9585401,zanonboito24_interspeech} as our speech encoder, fusing all hidden layers with learnable softmax weights, and inject frame-level audio features into the noise via cross-attention. Notably, in each DiT block, the self-attention layer runs over all $(H+T)$ motion tokens, while the audio cross-attention layer is applied only to the $T$ current-window tokens.  

To help the model interpret the audio source, we apply a state-conditioned FiLM modulation~\cite{perez2018film} to audio features before cross-attention:
\begin{equation}
\hat{\mathbf{A}}^{\tau}=\mathbf{A}^{\tau}\odot\big(1+\boldsymbol{\alpha}(q^{\tau})\big)+\boldsymbol{\beta}(q^{\tau}),
\end{equation}
where $[\boldsymbol{\alpha}(q^{\tau}),\boldsymbol{\beta}(q^{\tau})]\in\mathbb{R}^{2D}$
is produced by a lightweight zero-initialized MLP.
Each transformer block has its own modulation parameters, while parameters are shared across time steps within a block. For temporal alignment, we also use a diagonal locality mask in cross-attention: $M_{i,j}=0$ if $|i-j|\le R$, and $M_{i,j}=-\infty$ otherwise, where $R{=}2$ frames.

\subsubsection{Global conditioning via AdaLN.}
Following mainstream DiT paradigms~\cite{Peebles_2023_ICCV,chen2023pixartalphafasttrainingdiffusion}, we introduce global conditioning through AdaLN~\cite{yang2023diffusion}. Specifically, a unified global conditioning vector is constructed by concatenating: (i) the flow timestep embedding $\mathbf{t}$, (ii) a reference embedding from shape and reference motion ($[\mathbf{s},\mathbf{x}_{\mathrm{ref}}]$), (iii) a mean-pooled global audio embedding computed over current-window audio tokens only, using a separate set of learnable layer-fusion weights independent of those used in cross-attention, and (iv) the motion magnitude guidance $\mathbf{m}=[\mathbf{m}_\mathrm{r},\mathbf{m}_\mathrm{t}]\in[0,1]^2$.

To mitigate over-averaging caused by weak audio-motion correlation, we explicitly inject motion magnitude, which offers superior interpretability and practical utility over implicit style features~\cite{sunDiffPoseTalkSpeechDrivenStylistic2024,chuARTalkSpeechDriven3D2025}. During training, $\mathbf{m}_\mathrm{r}$, $\mathbf{m}_\mathrm{t}$ are computed from the ground-truth target window as the mean successive-frame rotational displacement (SO(3) geodesic angle) and translational displacement magnitude, and then normalized to $[0,1]$.

\subsection{Listening-Speaking State Conditioning}
\label{sec:ls_control}
As discussed in \cref{sec:intro}, dual-stream audio conditioning is ill-suited for causal user--LLM interaction because future user audio is unavailable. We therefore adopt a \emph{single-stream} conditioning interface and explicitly represent the behavioral mode with a per-frame listening-speaking state (LS-state). Training uses a unified single-stream waveform with aligned LS-state supervision. At inference time, a Streaming Audio Scheduler causally maps the microphone stream and the LLM stream to the same conditioning window and outputs the aligned LS-state.

\subsubsection{LS-state definition and acquisition.}
We define $q^{\tau}\in\{-1,+1\}$ with $+1$ for speaking and $-1$ for listening, and denote the window state as $\mathbf{q}\in\{-1,+1\}^{T}$. During training, we run TalkNet-ASD~\cite{10.1145/3474085.3475587} on each video to obtain a frame-level speaking confidence, apply a short temporal smoothing to suppress jitter, and then binarize it into $\mathbf{q}$. During inference, $\mathbf{q}$ is constructed from \emph{audio provenance}: frames sourced from LLM are labeled speaking ($+1$), while frames sourced from the environment are labeled listening ($-1$).

\subsubsection{LS-state injection.}
We inject $\mathbf{q}$ through two mechanisms described in \cref{sec:mc_dit}. We concatenate $q^{\tau}$ to each motion token in the history and current window. This provides an explicit mode indicator at the input representation and yields a mode-consistent temporal context. We apply a state-conditioned FiLM to frame-level audio features before cross-attention in each DiT block. This maintains the conditioning signal throughout the network depth.

\begin{algorithm}[t]
\caption{Streaming audio scheduler}
\label{alg:audio_buffer}
\small
\begin{algorithmic}[1]
\Require Listening queue $\mathcal{Q}_{\mathrm{L}}$, Speaking queue $\mathcal{Q}_{\mathrm{S}}$, cursors $(c_{\mathrm{L}}, c_{\mathrm{S}})$, previous mode $m_{\mathrm{prev}}$, window length $T$
\Ensure Waveform $\mathbf{a}$ and LS-state $\mathbf{q}\in\{-1,+1\}^{T}$
\State $U \gets \textsc{GetUnconsumedLength}(\mathcal{Q}_{\mathrm{S}}, c_{\mathrm{S}})$ \Comment{unconsumed frames in speaking queue}
\If{$U \ge T$} \Comment{speaking queue alone fills the window}
    \State $\mathbf{a} \gets \textsc{SliceHead}(\mathcal{Q}_{\mathrm{S}}, c_{\mathrm{S}}, T)$; \quad $\mathbf{q} \gets +\mathbf{1}_{T}$
    \State $c_{\mathrm{S}} \gets c_{\mathrm{S}} + T$; \quad $m_{\mathrm{prev}} \gets \text{speak}$
\ElsIf{$U > 0$} \Comment{partial Speak: fill remainder with listening tail}
    \State $x_{\mathrm{S}} \gets \textsc{SliceHead}(\mathcal{Q}_{\mathrm{S}}, c_{\mathrm{S}}, U)$; \quad $x_{\mathrm{L}} \gets \textsc{SliceTail}(\mathcal{Q}_{\mathrm{L}}, T - U)$
    \State $\mathbf{a} \gets (m_{\mathrm{prev}} = \text{speak})\ ?\ x_{\mathrm{S}} \Vert x_{\mathrm{L}} : x_{\mathrm{L}} \Vert x_{\mathrm{S}}$ \Comment{preserve temporal continuity}
    \State $\mathbf{q} \gets \textsc{LabelBySource}(\mathbf{a})$; \quad $c_{\mathrm{S}} \gets c_{\mathrm{S}} + U$
    \State $m_{\mathrm{prev}} \gets (m_{\mathrm{prev}} = \text{speak})\ ?\ \text{listen} : \text{speak}$ \Comment{track window tail state}
\Else \Comment{no pending Speak: pure listening mode}
    \State $\mathbf{a} \gets \textsc{SliceTail}(\mathcal{Q}_{\mathrm{L}}, T)$; \quad $\mathbf{q} \gets -\mathbf{1}_{T}$; \quad $m_{\mathrm{prev}} \gets \text{listen}$
\EndIf
\State \Return $(\mathbf{a}, \mathbf{q})$
\end{algorithmic}
\end{algorithm}

\subsubsection{Streaming Audio Scheduler.}
We maintain two audio queues: $\mathcal{Q}_{\mathrm{L}}$ as a rolling buffer (microphone) and $\mathcal{Q}_{\mathrm{S}}$ as a monotonically consumed queue. At each tick, we output a fixed-length window of $T$ frames by prioritizing unconsumed samples in $\mathcal{Q}_{\mathrm{S}}$ and filling any deficit with the most recent context from $\mathcal{Q}_{\mathrm{L}}$. When both sources appear within one window, we order the two segments using the previous window's tail mode $m_{\mathrm{prev}}$ to reduce boundary discontinuities; $\mathbf{q}$ is emitted alongside $\mathbf{a}$ by labeling each frame by its source (Alg.~\ref{alg:audio_buffer}).

\subsection{Two-Stage Training}
\label{sec:two_stage}
The FLAME parameters estimated from monocular videos are inevitably noisy and biased, especially in frames with occlusion, motion blur, or extreme pose, where tracker confidence is low. Pure coefficient-space supervision therefore forces the model to fit unreliable targets that are not always consistent with image evidence. We thus adopt a two-stage strategy: coefficient pretraining for stable dynamics, followed by end-to-end image refinement for bias correction.


\subsubsection{Stage I: Coefficient-space pretraining.}
We partition the per-frame motion vector $\mathbf{x}_1^{\tau}=[\mathbf{e}^{\tau},\,\mathbf{p}^{\tau}]$ into five parameter groups $\mathcal{K}$ (expression, jaw, eye, rotation, and translation) and optimize flow matching with per-group reweighting. The Stage-I objective is
\begin{equation}
\mathcal{L}_{\mathrm{I}}
= \sum_{k\in\mathcal{K}} \lambda_k
\left\|
\mathbf{v}_{\theta}^{(k)}-(\mathbf{x}_1^{(k)}-\mathbf{x}_0^{(k)})
\right\|_2^2
+ \lambda_{\mathrm{smooth}}\,\mathcal{L}_{\mathrm{smooth}}^{\mathrm{pose}}.
\label{eq:stage1_compact}
\end{equation}
$\mathcal{L}_{\mathrm{smooth}}^{\mathrm{pose}}=\sum_{\tau}\|\mathbf{p}^{\tau+1}-\mathbf{p}^{\tau}\|_2^2$ is applied only to rotation and translation dimensions to suppress jitter from video cuts and tracking instability; expression and jaw are excluded to avoid over-smoothing lip articulation.

\subsubsection{Stage II: End-to-end image refinement.}
For image-domain refinement, we use GAGAvatar~\cite{chu2024gagavatar_neurips} as the differentiable renderer. While it renders fewer facial details than high-fidelity avatar pipelines~\cite{qianGaussianAvatarsPhotorealisticHead2024}, its single-image avatar construction makes end-to-end fine-tuning practical on large-scale 2D video datasets. Moreover, GAGAvatar explicitly models intra-oral regions (\eg, teeth and tongue), which is crucial for realistic speech appearance. In Stage~II, we unfreeze the renderer and jointly optimize both the DiT and differentiable renderer parameters with image-domain losses, allowing the renderer to co-adapt to the generated motion distribution.

Using Rectified Flow, we can estimate the endpoint with a single Euler update,
\begin{equation}
\hat{\mathbf{x}}_1 \approx \mathbf{x}_t + (1-t)\,\mathbf{v}_\theta(\mathbf{x}_t,t,\mathbf{c}),
\label{eq:stage2_onestep}
\end{equation}
where $\mathbf{x}_t$ is sampled by Eq.~(\ref{eq:rf_path}). The straight path has a constant target velocity, so flow matching learns a displacement field; consequently, the inference step (setting $t{=}0$) coincides with the rendered endpoint, and image losses supervise exactly what will be sampled.

We then render the reconstructed endpoint with GAGAvatar $\hat{\mathbf{I}}=\mathcal{R}(\hat{\mathbf{x}}_1,\mathbf{s})$,
and apply image-domain losses to jointly refine both the motion model and the renderer,
\begin{equation}
\mathcal{L}_{\mathrm{II}}=
\lambda_{\mathrm{coef}}\,\mathcal{L}_{\mathrm{I}}+
\lambda_{\mathrm{img}}
\Big(\lambda_{1}\|\hat{\mathbf{I}}-\mathbf{I}\|_{1}+\lambda_{p}\|\psi(\hat{\mathbf{I}})-\psi(\mathbf{I})\|_2^2\Big),
\label{eq:stage2_loss}
\end{equation}
where $\psi(\cdot)$ is a frozen AlexNet-based LPIPS perceptual feature extractor~\cite{zhang2018lpips}.

\section{Experiments}

\subsection{Experimental Setup}

\subsubsection{Datasets.}
To ensure the robustness and generalization of our proposed model, we curated a diverse dataset comprising 11,372 samples, totaling 27.2 hours of footage. The primary data source is filtered from the TFHP dataset~\cite{sunDiffPoseTalkSpeechDrivenStylistic2024}, contributing 14.7 hours. To enhance linguistic and environmental diversity, we integrated 7.8 hours of data from the TalkVid~\cite{chen2025talkvid} and VFHQ~\cite{Xie_2022_CVPR} datasets. We additionally curated 3.9 hours of segments from the RealTalk dataset~\cite{geng2023affective} to specifically model listening behaviors. We extract 3D FLAME parameters from videos using the GAGAvatar~\cite{chu2024gagavatar_neurips} tracking pipeline. For evaluation, we randomly sample 100 speaking clips and 50 listening clips as the test set.

\subsubsection{Implementation Details.}
For the audio condition, we adopt mHuBERT-147~\cite{9585401,zanonboito24_interspeech} for multilingual adaptation. The motion branch predicts FLAME parameters over a 100-frame window, with 75 historical motion frames compressed into 20 frames using HCP. We train the model in two stages on two NVIDIA RTX 3090 GPUs: 80K iterations with lr=6e-4 and batch size 64, followed by 150K iterations with lr=8e-5 and batch size 4. During inference, we use 4-step Euler integration and set the motion-magnitude guidance to 0.3. More details are in the supplementary materials.




\subsubsection{Baselines.}
To comprehensively evaluate EmbodiedHead, we select state-of-the-art baselines across two distinct tracks. For standard speech-driven motion generation, we compare our method with FaceFormer~\cite{Fan_2022_CVPR}, CodeTalker~\cite{Xing_2023_CVPR}, DiffPoseTalk~\cite{sunDiffPoseTalkSpeechDrivenStylistic2024}, and ARTalk~\cite{chuARTalkSpeechDriven3D2025}. To explicitly assess interactive conversational dynamics, we benchmark against DualTalk~\cite{pengDualTalkDualSpeakerInteraction2025}, UniLS~\cite{chuUniLSEndtoEndAudioDriven2025}, and TIMAR~\cite{chenSeamlessInteractionCausal2025}, which are recent frameworks for unified listening-and-speaking motion generation. For visualization and image-domain evaluation, all baselines are rendered using the GAGAvatar~\cite{chu2024gagavatar_neurips}.



\begin{table}[t]
\centering
\caption{
Quantitative comparisons of 2D visual quality and 3D motion generation against baselines. The upper part reports on the EmbodiedHead speaking test set to evaluate standard speaking ability; BA is inapplicable to FaceFormer and CodeTalker because they do not predict head motion. The lower part reports on the public DualTalk test set to evaluate interactive listening-speaking ability; image-domain metrics are unavailable because the DualTalk test set does not provide ground-truth RGB frames.
}
\label{tab:main_results}
\small
\setlength{\tabcolsep}{4pt}
\renewcommand{\arraystretch}{1.05}
\begin{tabular}{lcccccccc}
\toprule
Method & LVE$\downarrow$ & FDD$\downarrow$ & MOD$\downarrow$ & BA$\uparrow$ & SID$\uparrow$ & PSNR$\uparrow$ & SSIM$\uparrow$ & LPIPS$\downarrow$ \\
\midrule
FaceFormer~\cite{Fan_2022_CVPR} & 11.02 & 3.46 & 3.84 & $-$ & 1.21 & 17.64 & 0.605 & 0.203 \\
CodeTalker~\cite{Xing_2023_CVPR} & 10.99 & 3.09 & 3.89 & $-$ & 1.32 & 17.68 & 0.607 & 0.202 \\
DiffPoseTalk~\cite{sunDiffPoseTalkSpeechDrivenStylistic2024} & 10.3 & 2.69 & 4.24 & 0.44 & 2.01 & 16.09 & 0.565 & 0.268 \\
ARTalk~\cite{chuARTalkSpeechDriven3D2025} & 6.86 & 2.42 & 3.21 & 0.46 & 2.39 & 16.81 & 0.579 & 0.221 \\
Ours (Stage1) & \textbf{5.70} & 1.66 & \textbf{2.98} & \textbf{0.47} & \textbf{2.64} & 17.53 & 0.600 & 0.199 \\
Ours (Stage2) & 5.71 & \textbf{1.58} & 3.04 & 0.46 & \textbf{2.64} & \textbf{18.28} & \textbf{0.617} & \textbf{0.184} \\
\midrule
DualTalk~\cite{pengDualTalkDualSpeakerInteraction2025} & 7.94 & 1.95 & 2.84 & \textbf{0.42} & 2.47 & $-$ & $-$ & $-$ \\
UniLS~\cite{chuUniLSEndtoEndAudioDriven2025} & 12.73 & 3.04 & 4.83 & 0.36 & 1.57 & $-$ & $-$ & $-$ \\
TIMAR~\cite{chenSeamlessInteractionCausal2025} & 8.02 & 1.94 & 2.98 & 0.41 & 2.57 & $-$ & $-$ & $-$ \\
Ours & \textbf{7.18} & \textbf{1.74} & \textbf{2.57} & 0.39 & \textbf{2.67} & $-$ & $-$ & $-$ \\
\bottomrule
\end{tabular}
\end{table}

\subsection{Quantitative Evaluation}

\subsubsection{Metrics.}
Our evaluation metrics are organized around the three core objectives of a head-embodied LLM avatar. For \emph{high visual quality}, we adopt standard image-domain metrics: PSNR, SSIM~\cite{wang2004ssim}, and LPIPS~\cite{zhang2018lpips}, which directly reflect the rendered fidelity perceived by users. For \emph{motion fidelity and listening-speaking behavior}, we use LVE~\cite{richard2021meshtalk} and MOD~\cite{Xing_2023_CVPR} for lip synchronization, FDD~\cite{Xing_2023_CVPR} for upper-face naturalness, BA~\cite{siyao2022bailando} for head-motion rhythmic synchronization, and SID~\cite{pengDualTalkDualSpeakerInteraction2025} for generation diversity. \emph{Real-time efficiency} is evaluated by the generation speed measured in frames per second (FPS).

\begin{figure}[t]
\centering
\includegraphics[width=11.5cm]{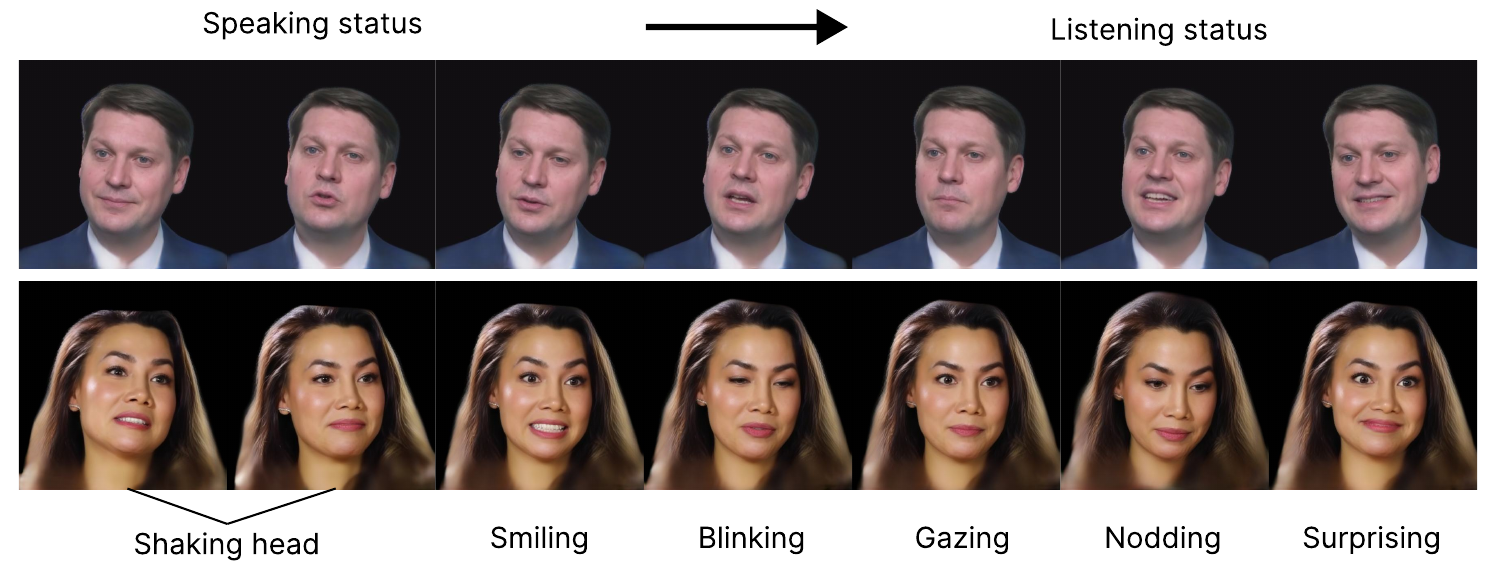}
\caption{Qualitative examples of natural listening-speaking transitions and conversational behaviors.}
\label{fig:qual_us}
\end{figure}

\subsubsection{2D Visual Quality Evaluation.}
A practical head-embodied LLM avatar must deliver high rendered visual quality to users. We evaluate the end-to-end rendered output by routing all methods through the identical render pipeline and comparing PSNR, SSIM, and LPIPS on the EmbodiedHead speaking test set (\cref{tab:main_results}, upper). As shown, EmbodiedHead achieves the best performance across all metrics, validating the efficacy of our DiT and two-stage training paradigm. Unlike previous methods reliant on noise-prone coefficient-space supervision, our Rectified Flow formulation enables direct one-step endpoint generation ($\hat{\mathbf{x}}_1$). This property seamlessly facilitates joint fine-tuning of the DiT and renderer in the second stage, effectively mitigating tracking biases, leading to superior 2D visual outcomes.

\subsubsection{3D Motion Evaluation.}
To assess motion fidelity, we evaluate the models under two distinct datasets: a standard speaking mode on the EmbodiedHead speaking test set, and an interactive listening-speaking mode on the DualTalk test set. Quantitative comparisons of 3D motion generation performance are summarized in \cref{tab:main_results}. In the standard speaking mode, EmbodiedHead significantly outperforms previous baselines, achieving state-of-the-art results across all evaluated metrics. The substantial improvements in LVE and FDD indicate that our model captures highly accurate lip articulations and natural upper-face dynamics to support realistic rendering. In the interactive listening-speaking mode, we train on the DualTalk train split and evaluate on its test set. Despite relying on a single audio stream with explicit LS-state conditioning (\cref{sec:ls_control}), EmbodiedHead surpasses the dual-audio-driven baseline methods in LVE, FDD, MOD, and SID. Although DualTalk and TIMAR attain slightly higher BA scores (0.42 and 0.41 vs.\ 0.39), this likely stems from their use of interlocutor visual cues to better capture conversational rhythm.



\subsubsection{Real-Time Performance.}
Real-time generation is a prerequisite for deploying a head-embodied LLM avatar in live conversation. Our full pipeline runs in real time with GAGAvatar as the renderer, achieving 59 FPS on a single NVIDIA RTX 3090 GPU. When measuring only the motion-coefficient generation module, our method reaches over 900 FPS, largely enabled by the efficiency of Rectified Flow. Compared with other methods, DiffPoseTalk is constrained by its DDPM architecture and cannot achieve real-time performance. Although some baselines can reach real-time speeds at the motion-coefficient level, none of them explores end-to-end image generation, so their reported efficiency does not reflect the actual speed of producing the final rendered output.

\begin{figure}[t]
\centering
\includegraphics[width=11.4cm]{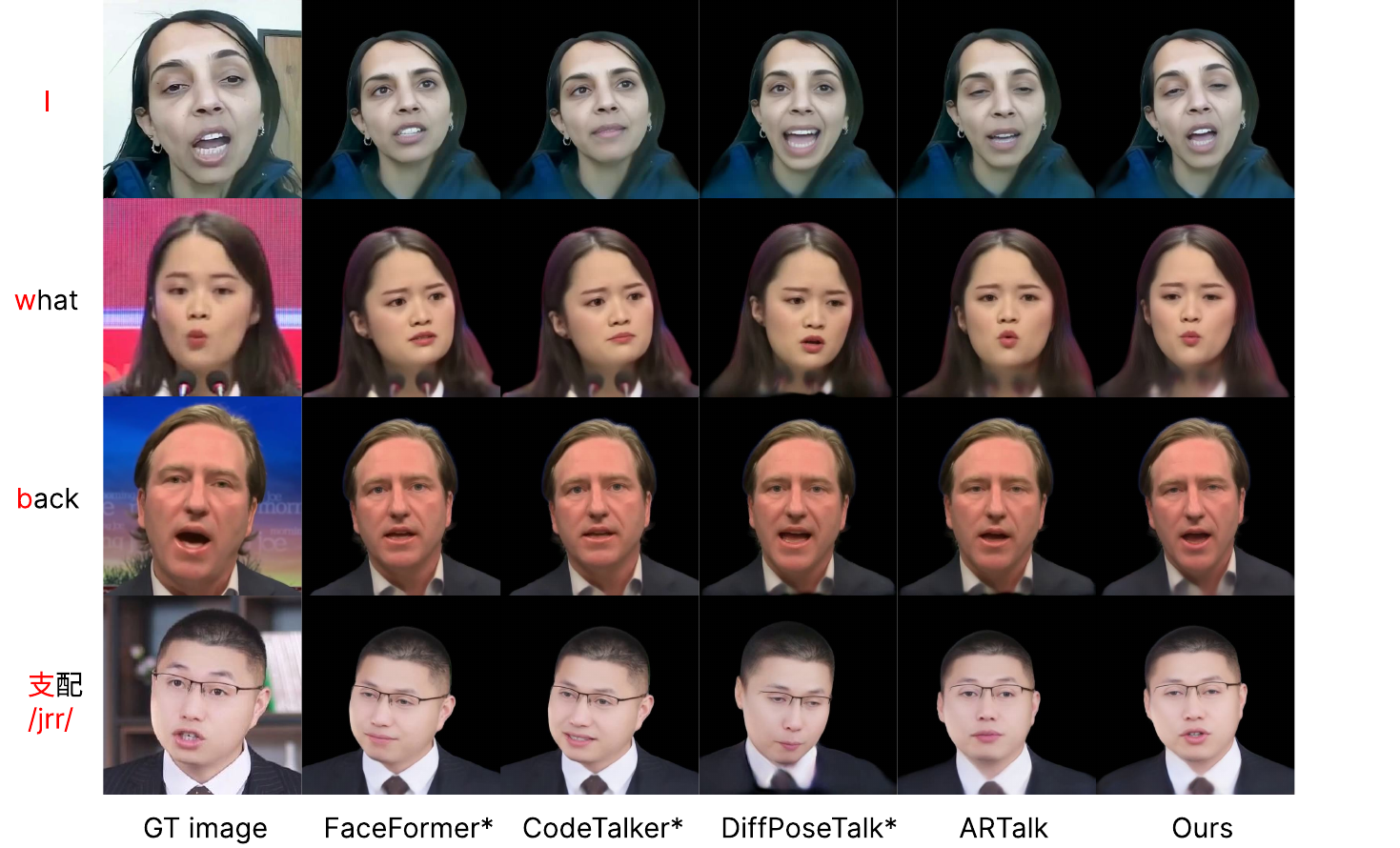}
\caption{Qualitative comparisons of 2D visual quality against other baseline methods on the EmbodiedHead speaking test set. * indicates baselines are additionally rendered.}
\label{fig:qual_2d}
\end{figure}

\begin{figure}[!htbp]
\centering
\includegraphics[width=10.6cm]{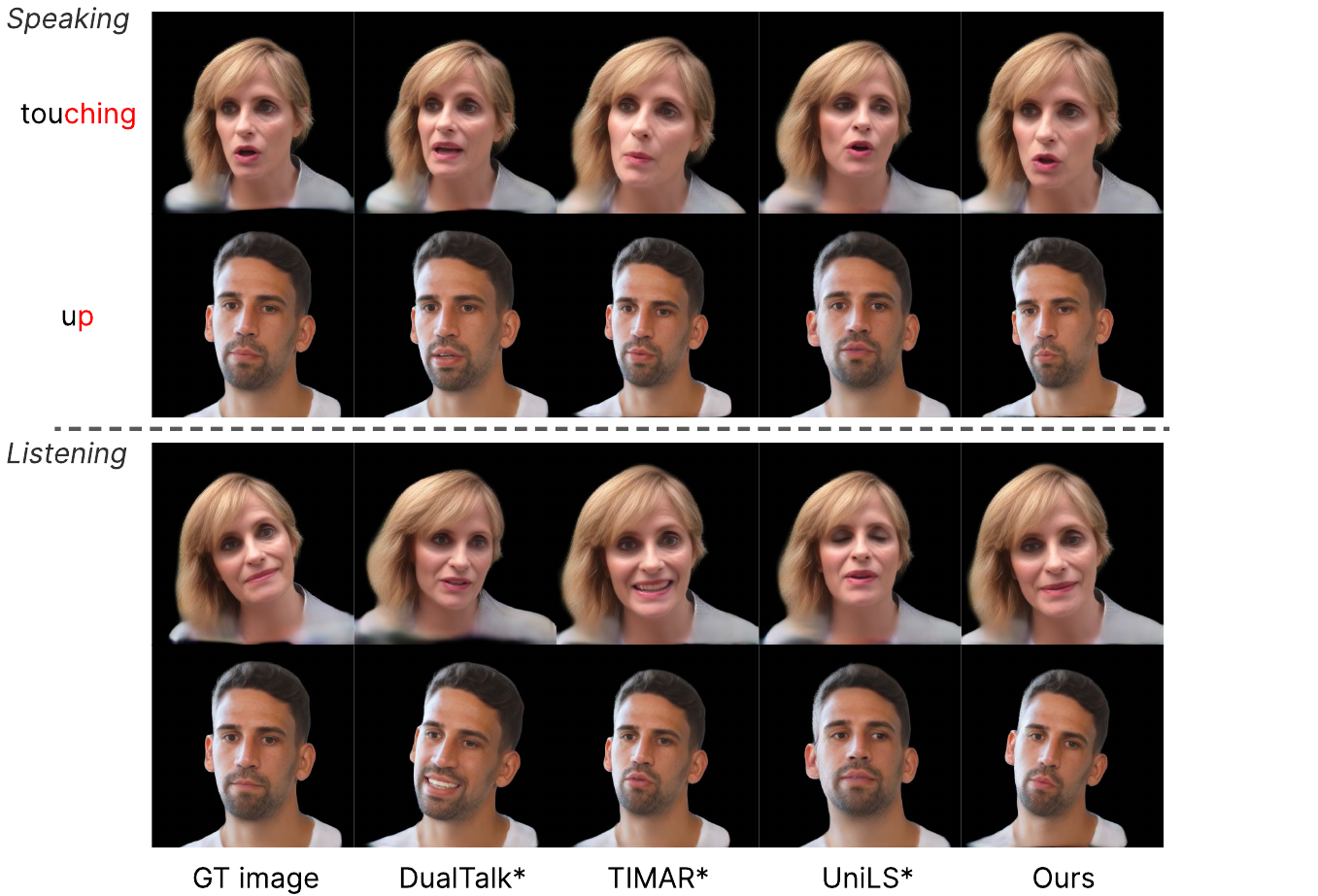}
\caption{Qualitative comparisons of 2D visual quality against interactive baseline methods on the DualTalk test set, covering both speaking (upper) and listening (lower) statuses. * indicates baselines are additionally rendered.}
\label{fig:qual_dual}
\end{figure}

\begin{table}[ht]
\centering
\caption{User study results. Each value reports the percentage of all pairwise comparisons in which participants preferred our method over the corresponding baseline, with ties included.}
\label{tab:user_study}
\small
\setlength{\tabcolsep}{6pt}
\renewcommand{\arraystretch}{1.1}
\begin{tabular}{@{}r@{\ vs.\ }lcccc@{}}
\toprule
\multicolumn{2}{c}{Comparison}
& \shortstack{Overall\\Realism}
& \shortstack{Lip-Sync\\Accuracy}
& \shortstack{Listening\\Naturalness}
& \shortstack{Turn-Taking\\Smoothness} \\
\midrule
Ours & DualTalk~\cite{pengDualTalkDualSpeakerInteraction2025} & 89.1\% & 81.8\% & 85.5\% & 85.0\% \\
Ours & TIMAR~\cite{chenSeamlessInteractionCausal2025}    & 91.8\% & 90.9\% & 86.8\% & 87.7\% \\
Ours & UniLS~\cite{chuUniLSEndtoEndAudioDriven2025}    & 76.8\% & 72.7\% & 69.5\% & 68.6\% \\
\bottomrule
\end{tabular}
\end{table}

\subsection{Qualitative Evaluation}

To demonstrate our interactive capabilities, \cref{fig:qual_us} visualizes the natural listening-speaking transitions and conversational behaviors generated by EmbodiedHead. Modulated by the explicit LS-state, our model achieves seamless state switching without abrupt visual artifacts. Beyond accurate lip synchronization, the rendered 2D avatars exhibit rich, context-aware non-verbal dynamics, such as attentive nodding and blinking during listening, alongside vivid facial expressions during speaking. These high-fidelity interactive behaviors significantly enhance the realism and engagement of the conversational agent.

In standard speaking scenarios (\cref{fig:qual_2d}), FaceFormer, CodeTalker, and ARTalk often yield restricted mouth amplitudes, failing to fully articulate phonetic details. Although DiffPoseTalk exhibits larger motion amplitudes, it still suffers from noticeable lip-audio desynchronization. In conversational settings (\cref{fig:qual_dual}), despite producing rich dynamics, interactive baselines occasionally hallucinate unreasonable mouth openings during the ``listening'' phase. By contrast, equipped with an explicit state modulation mechanism, EmbodiedHead effectively suppresses unnatural mouth hallucinations when listening, while ensuring highly accurate and expressive lip synchronization when speaking. 

We conducted a randomized pairwise preference study to evaluate the perceptual quality of the generated conversational avatars. Our user study focuses on interactive listening-speaking baselines, including DualTalk~\cite{pengDualTalkDualSpeakerInteraction2025}, TIMAR~\cite{chenSeamlessInteractionCausal2025}, and UniLS~\cite{chuUniLSEndtoEndAudioDriven2025}, since these methods generate interactive behaviors whose naturalness is best assessed by human judgment. The study involved 22 voluntary participants, each evaluating 48 pairwise comparisons across speaking and listening scenarios using four criteria: Overall Realism, Lip-Sync Accuracy, Listening Naturalness, and Turn-Taking Smoothness. For each comparison, participants selected the better result or indicated a tie. As shown in Tab.~\ref{tab:user_study}, our method is consistently preferred over all baselines across the four criteria, with preference rates of at least 68.6\%, demonstrating its perceptual advantages in conversational listening-speaking interaction. Further details are provided in the supplementary material.

\begin{table}[t]
\centering
\caption{Ablations on global and listening modules. Upper: global condition ablation; lower: listening-speaking module ablation. \textit{mag}, \textit{ref}, and \textit{audio} denote motion magnitude, reference embedding, and audio embedding, respectively.}
\label{tab:ablation-module}
\small
\setlength{\tabcolsep}{5pt} 
\renewcommand{\arraystretch}{1.1}
\begin{tabular}{lcccc}
\toprule
Method & LVE$\downarrow$ & FDD$\downarrow$ & MOD$\downarrow$ & BA$\uparrow$\\
\midrule
Baseline           & 5.97          & 1.70          & 2.89          & 0.41          \\
\quad +mag        & 6.20          & 1.58          & 2.96          & 0.40          \\
\quad +mag+ref        & 5.84          & 1.63          & 2.83          & 0.42          \\
\quad +mag+ref+audio & \textbf{5.76} & \textbf{1.55} & \textbf{2.63} & \textbf{0.43} \\
\midrule
Baseline           & 6.58          & 1.53          & 2.43          & 0.37          \\
\quad+LS-state      & 6.21          & \textbf{1.46} & 2.10          & 0.36          \\
\quad+LS-state+FiLM & \textbf{6.13} & \textbf{1.46} & \textbf{1.99} & \textbf{0.38} \\
\bottomrule
\end{tabular}
\end{table}

\begin{figure}[t]
\centering
\includegraphics[width=11.4cm]{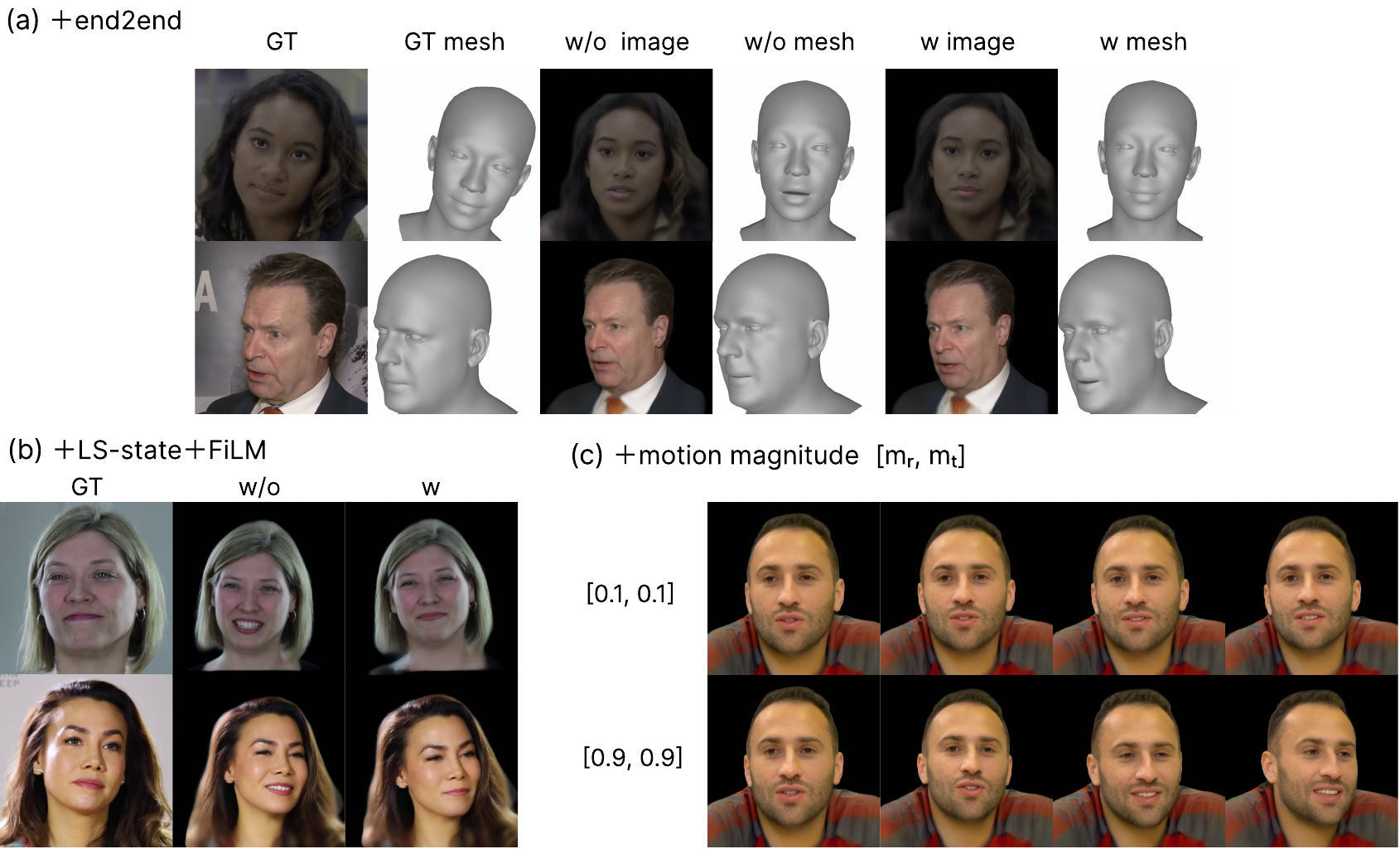}
\caption{Visual results of ablation studies. (a) The second-stage image-domain supervision improves both rendered quality and mesh geometry. (b) LS-state with FiLM modulation suppresses unnatural mouth openings during listening. (c) The motion magnitude condition controls head-motion amplitude under the same audio input.}
\label{fig:ablation}
\end{figure}

\subsection{Ablation Study}
\subsubsection{Effect of the Global Condition.}
We validate the global condition module in the top half of \cref{tab:ablation-module}. The baseline only uses the timestep condition. Progressively adding reference and audio embeddings to AdaLN steadily improves all metrics. The motion magnitude condition is defined as $\mathbf{m}=[m_r,m_t]$, where $m_r$ and $m_t$ denote the normalized rotation and translation amplitudes, respectively. As shown in \cref{fig:ablation}(c), small motion magnitude values (\eg, 0.1) yield restrained head motions while large values (\eg, 0.9) produce markedly more expressive movements, all from the same audio input. This continuous controllability demonstrates effective disentanglement of motion amplitude from the driving signal, enhancing both diversity and user-level customizability.


\begin{table}[t]
\centering
\caption{Ablation on inference-step scheduling.}
\label{tab:ablation-step}
\small
\setlength{\tabcolsep}{5pt} 
\renewcommand{\arraystretch}{1.1}
\begin{tabular}{lccccccc}
\toprule
Step & LVE$\downarrow$ & FDD$\downarrow$ & MOD$\downarrow$ & BA$\uparrow$ & PSNR$\uparrow$ & SSIM$\uparrow$ & LPIPS$\downarrow$ \\
\midrule
1     & 5.80  & 1.76  & 2.73  & \textbf{0.48} & 17.08 & 0.571 & 0.204        \\
4     & \textbf{5.76}  & 1.55  & \textbf{2.63}  & 0.43 & \textbf{17.29} & \textbf{0.578} & \textbf{0.202}        \\
10    & 5.99  & 1.46  & 2.67  & 0.43 & 17.13 & 0.573 & 0.206        \\
25    & 6.18  & \textbf{1.44}  & 2.73  & 0.41 & 17.10 & 0.572 & 0.208        \\
\bottomrule
\end{tabular}
\\[2pt]
\end{table}

\subsubsection{Effect of the Listening-Speaking Module.}
We assess explicit LS-state injection in the bottom half of \cref{tab:ablation-module}, which is the component most directly tied to our unified listening-speaking objective. When LS-state conditioning is removed, the model attempts to infer conversational states directly from the audio signal. As illustrated in \cref{fig:ablation}(b), it tends to interpret weak or distant speech as a listening state, and under uncertain conditions may open its mouth without producing clear speaking motions. Adding the state to the model input reduces MOD from 2.43 to 2.10, and further applying FiLM modulation to cross-attention brings it to 1.99, effectively eliminating spurious articulatory artifacts during listening and enabling reliable state-dependent behavior.

\subsubsection{Effect of the Two-Stage Training.}
We evaluate the impact of the second-stage image-domain supervision. As shown in \cref{tab:main_results}(upper) and \cref{fig:ablation}(a), this strategy significantly enhances 2D visual fidelity. Furthermore, it effectively mitigates inherent tracking noise. For instance, as depicted in the second row of \cref{fig:ablation}(a), our optimized geometry successfully captures subtle, natural mouth openings, visually surpassing the tracked pseudo-GT FLAME mesh. This corroborates our hypothesis that 2D supervision can effectively rectify 3D tracker biases.

\subsubsection{Effect of the Inference Step.}
Our inference step analysis (\cref{tab:ablation-step}) demonstrates that the proposed architecture enables high-quality one-step generation. Specifically, 1-step inference yields performance highly comparable to the 25-step setting. For additional details and ablation results, please refer to the supplementary material.

\section{Conclusion}
We presented EmbodiedHead, an end-to-end speech-driven talking-head framework for head-embodied LLM avatars. By coupling a Rectified-Flow DiT with a differentiable renderer, our pipeline achieves 59\,FPS on a single GPU with only four sampling steps, satisfying the real-time requirement for live conversation. Explicit per-frame listening-speaking state conditioning and a Streaming Audio Scheduler replace the dual-stream paradigm, enabling unified listening and speaking behavior without interlocutor look-ahead. A two-stage training strategy that augments coefficient-space flow matching with image-domain refinement further bridges the gap between mesh-level accuracy and rendered visual fidelity. Experiments on both our curated dataset and public benchmarks validate state-of-the-art performance across motion, rendering, and interaction metrics.

\section*{Acknowledgements}
This work was supported by the National Natural Science Foundation of China (62572194). This article was also supported by the Key Technology Research and Development Program Project of the Shanghai Science and Technology Commission under Grants (25511107200).

\clearpage

\bibliographystyle{splncs04}
\bibliography{main,3d_virtual_human}

\clearpage

\appendix

\begin{center}
    \Large\textbf{EmbodiedHead: Real-Time Listening and Speaking Avatar for Conversational Agents} \\
    \vspace{0.4cm}
    \large Supplementary Material
\end{center}
\vspace{0.5cm}

\setcounter{section}{0}
\renewcommand{\thesection}{\Alph{section}}
\renewcommand{\thesubsection}{\Alph{section}.\arabic{subsection}}

\makeatletter
\renewcommand{\theHsection}{supp.section.\arabic{section}}
\renewcommand{\theHsubsection}{supp.section.\arabic{section}.\arabic{subsection}}
\renewcommand{\theHfigure}{supp.figure.\arabic{figure}}
\renewcommand{\theHtable}{supp.table.\arabic{table}}
\renewcommand{\theHalgorithm}{supp.algorithm.\arabic{algorithm}}
\makeatother

\setcounter{table}{0}
\renewcommand{\thetable}{S\arabic{table}}
\setcounter{figure}{0}
\renewcommand{\thefigure}{S\arabic{figure}}

This supplementary material provides additional details to support the main manuscript. \cref{sec:supp_arch} details the model architecture. \cref{sec:supp_metrics} describes the hyperparameters and formalizes the evaluation metrics used in our experiments. \cref{sec:supp_exp} presents additional experimental results, including additional ablation results, the user study, and runtime analysis. \cref{sec:supp_ethics} discusses ethical considerations. \cref{sec:supp_limitations} discusses the limitations of our current framework and outlines directions for future research. For more qualitative results, please refer to the demos on our project page.


\section{Architecture Details}
\label{sec:supp_arch}
In the configuration used in our experiments, the Rectified-Flow~\cite{liu2022flowstraightfastlearning} DiT~\cite{Peebles_2023_ICCV,chen2023pixartalphafasttrainingdiffusion} predicts FLAME~\cite{FLAME:SiggraphAsia2017} motion coefficients directly. Each target frame contains 100 expression coefficients together with jaw pose, eye pose, global rotation, and global translation, giving a 115-dimensional motion vector in total. The model predicts a 100-frame target window at 25 FPS. We append a one-dimensional listening-speaking state to every motion frame before tokenization, so the per-token input dimension becomes 116. A linear projection maps each token to a 448-dimensional transformer~\cite{NIPS2017_3f5ee243} space. The DiT backbone uses 8 blocks, 8 attention heads, and an MLP ratio of 4, yielding a head dimension of 56 and a feed-forward hidden width of 1792.

For temporal context, the model consumes 75 historical frames in addition to the current noisy window. Following our History Context Packing (HCP) design, these 75 frames are split into four groups with receptive spans of 5, 10, 20, and 40 frames, respectively. Each group is compressed into 5 tokens by an independent linear projector, producing 20 packed history tokens in total. These 20 tokens are concatenated with the 100 current-window tokens, so self-attention is performed over a sequence of 120 motion tokens. The reference condition is formed by concatenating the first 100 FLAME shape coefficients with one reference motion frame, resulting in a 215-dimensional vector.

We use frozen mHuBERT-147~\cite{9585401,zanonboito24_interspeech} as the audio encoder. All hidden layers are fused by learnable softmax weights, with separate weight sets for the local cross-attention branch and the global conditioning branch. The extracted audio sequence is aligned to the motion rate by a strided 1D convolution with kernel size 5 and stride 2, followed by LayerNorm~\cite{ba2016layer}, producing 448-dimensional audio tokens. In cross-attention, we further apply a diagonal locality mask with radius 2 frames, so each motion token attends only to a narrow temporal neighborhood in the aligned audio sequence.

Global conditions are injected through AdaLN~\cite{yang2023diffusion}. The conditioning vector has dimension 896 and is formed by concatenating four parts: a 448-dimensional sinusoidal timestep embedding, a 112-dimensional reference embedding, a 224-dimensional mean-pooled global audio embedding, and a 112-dimensional motion magnitude embedding derived from the normalized rotation and translation guidance scalars. Each DiT block consists of RoPE self-attention, RoPE audio cross-attention, and a feed-forward layer, with rotary base $\theta=10000$~\cite{su2024roformer}. Before audio cross-attention, the aligned audio features are modulated by a block-specific FiLM~\cite{perez2018film} network conditioned on the listening-speaking state; this modulator uses a hidden size of 112 and zero-initialized output weights so that training starts from an identity mapping. The AdaLN modulation layers are also zero-initialized for stable optimization. Cross-attention is applied only to the 100 current-window tokens, while the packed history tokens contribute through self-attention only. Finally, a last AdaLN-modulated projection maps the final 100 tokens back to 115-dimensional velocity predictions.

\section{Experimental Setup and Metric Formalization}
\label{sec:supp_metrics}
\subsection{Training Hyperparameters and Loss Weights}

Both training stages utilize the AdamW~\cite{loshchilov2017decoupled} optimizer. For the flow-matching objective (Equation (6) of the main manuscript), the parameter group weights are empirically set to: $\lambda_{\mathrm{expr}}=0.6$, $\lambda_{\mathrm{jaw}}=0.1$, $\lambda_{\mathrm{eye}}=0.1$, $\lambda_{\mathrm{rot}}=0.1$, and $\lambda_{\mathrm{trans}}=0.1$. The pose smoothing weight is set to $\lambda_{\mathrm{smooth}}=0.1$.

For Stage I, we employ a linear learning rate warmup for the first 2,000 iterations and train with a total batch size of 64. During the Stage II end-to-end refinement, we employ 1,000 warmup steps and render the images at a resolution of 256$\times$256. The loss weights (Equation (8) of the main manuscript) are set as follows: the coefficient-space flow-matching weight is set to $\lambda_{\mathrm{coef}}=0.2$, while the image-domain L1 and perceptual loss weights are set to $\lambda_{1}=0.2$ and $\lambda_{p}=0.2$, respectively. We also scale the learning rate of the GAGAvatar~\cite{chu2024gagavatar_neurips} renderer by a factor of 0.5 relative to the base learning rate. Due to the memory footprint of the differentiable renderer, the total batch size in Stage II is reduced to 4.

\subsection{Metric Calculation Details}
To comprehensively evaluate both the 3D motion fidelity and the 2D visual quality, our metrics (LVE~\cite{richard2021meshtalk}, FDD~\cite{Xing_2023_CVPR}, MOD~\cite{Xing_2023_CVPR}, BA~\cite{siyao2022bailando}, SID~\cite{pengDualTalkDualSpeakerInteraction2025}, PSNR, SSIM~\cite{wang2004ssim}, and LPIPS~\cite{zhang2018lpips}) span geometric, temporal, and image domains. For 3D geometry-based metrics, the predicted FLAME~\cite{FLAME:SiggraphAsia2017} coefficients are mapped to mesh vertices $\mathbf{v}$ and 3D landmarks $\mathbf{p}$. In alignment with established evaluation protocols, we formalize the metrics as follows.



\subsubsection{Lip Vertex Error (LVE $\downarrow$).}
LVE~\cite{richard2021meshtalk} evaluates temporal lip synchronization by computing the maximum $L_2$ error across all vertices in the lip region ($V_{\text{lip}}$) for each frame $t$, and averaging over the sequence of length $T$:
\begin{equation}
\text{LVE} = \frac{1}{T} \sum_{t=1}^{T} \max_{i \in V_{\text{lip}}} \left\| \mathbf{v}_{t, i}^{\text{pred}} - \mathbf{v}_{t, i}^{\text{gt}} \right\|_2
\end{equation}
where $\mathbf{v}_{t, i} \in \mathbb{R}^3$ denotes the 3D coordinate of vertex $i$ at frame $t$.

\subsubsection{Face Dynamics Deviation (FDD $\downarrow$).}
FDD~\cite{Xing_2023_CVPR} assesses the naturalness of upper-face motions. Let $V_{\text{upper}}$ denote the subset of FLAME~\cite{FLAME:SiggraphAsia2017} mesh vertices corresponding to the eye and forehead regions. We first compute the relative vertex motion to the initial frame $\mathbf{m}_{t, i} = \mathbf{v}_{t, i} - \mathbf{v}_{1, i} \in \mathbb{R}^3$ to remove static offsets. We then measure the difference in the temporal standard deviation ($\sigma_i$) between the prediction and ground truth:
\begin{equation}
\sigma_i = \sqrt{\frac{1}{T}\sum_{t=1}^T \left\| \mathbf{m}_{t, i} - \bar{\mathbf{m}}_i \right\|_2^2}
\end{equation}
where $\bar{\mathbf{m}}_i = \frac{1}{T}\sum_{t=1}^T \mathbf{m}_{t, i}$ is the temporal mean. The FDD is formulated as:
\begin{equation}
\text{FDD} = \frac{1}{|V_{\text{upper}}|} \sum_{i \in V_{\text{upper}}} \left| \sigma_i^{\text{pred}} - \sigma_i^{\text{gt}} \right|
\end{equation}

\subsubsection{Mouth Opening Difference (MOD $\downarrow$).}
MOD~\cite{Xing_2023_CVPR} focuses on the similarity of mouth opening amplitudes. Let $d_t = \|\mathbf{p}_{t, 66} - \mathbf{p}_{t, 62}\|_2$ be the distance between the upper (idx: 62) and lower (idx: 66) lip landmarks. MOD is the mean absolute error of this distance:
\begin{equation}
\text{MOD} = \frac{1}{T} \sum_{t=1}^{T} \left| d_t^{\text{pred}} - d_t^{\text{gt}} \right|
\end{equation}

\subsubsection{Beat Alignment (BA $\uparrow$).}
BA~\cite{siyao2022bailando} quantifies the rhythmic synchronization of head movements. We extract the angular velocity of head rotation and detect motion peaks (beats) $B$ using a dynamic threshold. The alignment score is calculated symmetrically using a Gaussian kernel with $\sigma = 0.1\text{s}$:
\begin{equation}
\begin{aligned}
\text{BA} = \frac{1}{2} \Bigg( &\frac{1}{|B^{\text{pred}}|} \sum_{b \in B^{\text{pred}}} \exp\Big(-\frac{\min_{b' \in B^{\text{gt}}}(b - b')^2}{2\sigma^2}\Big) \\
&+ \frac{1}{|B^{\text{gt}}|} \sum_{b' \in B^{\text{gt}}} \exp\Big(-\frac{\min_{b \in B^{\text{pred}}}(b' - b)^2}{2\sigma^2}\Big) \Bigg)
\end{aligned}
\end{equation}
where $b, b' \in \mathbb{R}$ denote the timestamps of the detected beats in seconds.

\subsubsection{Diversity Metric (SID $\uparrow$).}
To measure the generation diversity from the same audio input, we use the Speaker Identity Diversity (SID)~\cite{pengDualTalkDualSpeakerInteraction2025}. We fit a $K$-Means clustering model on the ground truth feature distribution and predict the cluster assignments for the generated sequences. The diversity for a specific component $k$ is defined as the Shannon entropy of its cluster assignment histogram:
\begin{equation}
\text{SID}_{k} = -\sum_{c=1}^{K} p_c \log_2(p_c + \epsilon)
\end{equation}
where $p_c = n_c / N$ is the probability of a sample falling into cluster $c$, and $\epsilon = 10^{-8}$ is a small constant for numerical stability. For a comprehensive assessment, we compute the mean of the SID across three semantic groups: jaw ($\text{SID}_{\text{jaw}}$), the first 50 dominant dimensions of expression ($\text{SID}_{\text{exp50}}$), and rotation ($\text{SID}_{\text{rot}}$):
\begin{equation}
\text{SID} = \frac{1}{3} \left( \text{SID}_{\text{jaw}} + \text{SID}_{\text{exp50}} + \text{SID}_{\text{rot}} \right)
\end{equation}

\subsubsection{2D Metrics (PSNR $\uparrow$, SSIM $\uparrow$, LPIPS $\downarrow$).}
To evaluate the final visual quality, we synthesize 2D image frames by rendering the FLAME-driven 3D mesh via the GAGAvatar~\cite{chu2024gagavatar_neurips} pipeline. The rendered images are centrally cropped to a $256 \times 256$ region of interest (ROI) surrounding the face.
\begin{itemize}
    \item \textbf{PSNR \& SSIM:} We compute the Peak Signal-to-Noise Ratio (PSNR) and Structural Similarity Index (SSIM)~\cite{wang2004ssim} frame-by-frame between the rendered outputs and the preprocessed ground-truth video, then average across the temporal dimension. SSIM is evaluated using an $11 \times 11$ Gaussian window with $\sigma=1.5$.
    \item \textbf{LPIPS (AlexNet):} We adopt the AlexNet-based LPIPS~\cite{zhang2018lpips} implementation to measure perceptual similarity, reporting the average distance across all frames.
\end{itemize}

\section{Additional Experimental Results}
\label{sec:supp_exp}

\subsection{Additional Quantitative Ablation Results}
\label{sec:supp_ablation}

In this section, we provide more comprehensive quantitative ablation results to complement the main manuscript, specifically detailing the impact of inference-step scheduling and motion magnitude guidance.

\subsubsection{Ablation on Inference-Step Scheduling.}
We first present the complete ablation study on the inference steps in \cref{tab:supp_ablation_step}. The results demonstrate that our Rectified-Flow DiT architecture enables high-quality generation with only a few sampling steps. Specifically, 1-step inference already achieves performance comparable to the 25-step setting on most 3D motion and 2D image metrics. Weighing these findings, we ultimately select 4 steps as our default inference configuration to achieve an optimal balance between real-time speed, visual quality, and motion fidelity.

To further validate the mathematical grounding of our Stage-II image-domain fine-tuning, we report a special ``$1^\dagger$'' setting. This setting denotes one-step generation using randomly sampled timesteps that are strictly consistent with the training phase distribution. The performance of this practical one-step output supports the validity of our Stage-II optimization, rendering the end-to-end image-domain refinement both computationally feasible and highly efficient.

\begin{table}[ht]
\centering
\caption{Detailed ablation on inference-step scheduling. Best results are highlighted in \textbf{bold} except for $1^\dagger$. $1^\dagger$ denotes one-step generation with randomly sampled timesteps (training-consistent); results support the validity of Stage-II image-domain fine-tuning from practical one-step outputs.}
\label{tab:supp_ablation_step}
\small
\setlength{\tabcolsep}{6pt} 
\renewcommand{\arraystretch}{1.1}
\begin{tabular}{lccccccc}
\toprule
Step & LVE$\downarrow$ & FDD$\downarrow$ & MOD$\downarrow$ & BA$\uparrow$ & PSNR$\uparrow$ & SSIM$\uparrow$ & LPIPS$\downarrow$ \\
\midrule
1          & 5.80 & 1.76 & 2.73 & \textbf{0.48} & 17.08 & 0.571 & 0.204 \\
$1^\dagger$& 4.45 & 0.85 & 2.44 & 0.47 & 17.09 & 0.576 & 0.193 \\
2          & 5.79 & 1.74 & 2.67 & 0.45 & 17.23 & \textbf{0.578} & 0.206 \\
3          & 5.91 & 1.60 & 2.68 & 0.44 & 17.27 & \textbf{0.578} & \textbf{0.202} \\
4          & \textbf{5.76} & 1.55 & \textbf{2.63} & 0.43 & \textbf{17.29} & \textbf{0.578} & \textbf{0.202} \\
5          & 5.95 & 1.53 & 2.69 & 0.42 & 17.22 & 0.576 & \textbf{0.202} \\
10         & 5.99 & 1.46 & 2.67 & 0.43 & 17.13 & 0.573 & 0.206 \\
15         & 6.12 & 1.47 & 2.73 & 0.42 & 17.09 & 0.571 & 0.208 \\
25         & 6.18 & \textbf{1.44} & 2.73 & 0.41 & 17.10 & 0.572 & 0.208 \\
\bottomrule
\multicolumn{8}{l}{\footnotesize $^\dagger$ \textit{Not directly comparable to standard fixed-step inference settings.}}
\end{tabular}
\end{table}

\subsubsection{Ablation on Motion Magnitude.}
We quantitatively evaluate the explicit controllability introduced by our global condition module. As visualized in the main manuscript, modulating the motion magnitude values ($\mathbf{m} = [\mathbf{m}_\mathrm{r}, \mathbf{m}_\mathrm{t}]$) affords intuitive control over the kinematic intensity of the generated head dynamics. \cref{tab:supp_ablation_mag_matrix} details how varying these motion magnitude values influences the actual generated rotation and translation magnitudes. Notably, the results exhibit a degree of disentanglement: altering the rotation condition ($\mathbf{m}_\mathrm{r}$) primarily scales the rotational amplitude while leaving the translation magnitude mostly stable, and vice versa. For reference, the original Ground Truth (GT) sequences exhibit a mean rotation magnitude of 0.61 and a translation magnitude of 0.81. This baseline serves as a standard anchor, demonstrating that our model can effectively control the motion intensity, capable of generating both subtle, restrained motions (when setting small motion magnitude values, \eg, $0.1$) and highly expressive, larger-amplitude dynamics (when setting large motion magnitude values, \eg, $0.9$).

\begin{table}[ht]
\centering
\caption{Quantitative ablation on motion magnitude guidance. We report the actual generated \textit{Rotation} (Rot.) and \textit{Translation} (Trans.) magnitudes under different combinations of motion magnitude values $[\mathbf{m}_\mathrm{r}, \mathbf{m}_\mathrm{t}]$. The results demonstrate that our model not only effectively controls the magnitude but also partially disentangles rotation and translation.}
\label{tab:supp_ablation_mag_matrix}
\resizebox{0.95\textwidth}{!}{
\renewcommand{\arraystretch}{1.3}
\begin{tabular}{c | cc | cc | cc}
\hline
\multirow{3}{*}{\begin{tabular}{@{}c@{}}\textbf{Input Trans.} \\ \textbf{($\mathbf{m}_\mathrm{t}$)}\end{tabular}} & \multicolumn{6}{c}{\textbf{Input Rot. ($\mathbf{m}_\mathrm{r}$)}} \\
\cline{2-7}
 & \multicolumn{2}{c|}{\textbf{0.1}} & \multicolumn{2}{c|}{\textbf{0.5}} & \multicolumn{2}{c}{\textbf{0.9}} \\
\cline{2-7}
 & \makebox[1.4cm]{\textit{Rot.}} & \makebox[1.4cm]{\textit{Trans.}} & \makebox[1.4cm]{\textit{Rot.}} & \makebox[1.4cm]{\textit{Trans.}} & \makebox[1.4cm]{\textit{Rot.}} & \makebox[1.4cm]{\textit{Trans.}} \\
\hline
\textbf{0.1} & 0.50 & 0.88 & 1.05 & 0.95 & 1.77 & 1.07 \\
\textbf{0.5} & 0.55 & 1.41 & 1.03 & 1.56 & 1.64 & 1.72 \\
\textbf{0.9} & 0.76 & 2.57 & 1.10 & 3.05 & 1.31 & 2.63 \\
\hline
\textbf{GT} & \multicolumn{6}{c}{\textit{Rot.}: 0.61 \qquad\qquad\qquad \textit{Trans.}: 0.81} \\
\hline
\end{tabular}
}
\end{table}

\subsection{User Study}
\label{sec:supp_userstudy}

Fig.~\ref{fig:user_study} shows the interface used in our user study. Before starting the comparisons, participants were presented with the study instructions, an audio playback check, and a consent statement for voluntary participation. During the evaluation, two anonymized videos were shown side by side for each comparison, with randomized left-right order. Participants then selected the preferred result or indicated a tie according to the four evaluation criteria: Overall Realism, Lip-Sync Accuracy, Listening Naturalness, and Turn-Taking Smoothness.

\begin{figure}[t]
\centering
\includegraphics[width=\linewidth]{user_study.pdf}
\caption{Screenshot of the user study interface. The left panel shows the study instructions, audio playback check, and consent statement. The right panel shows the randomized pairwise comparison interface with four evaluation criteria.}
\label{fig:user_study}
\end{figure}

\subsection{Runtime and Latency Analysis}
\label{sec:supp_latency}
We profile the runtime of the main components on a single NVIDIA RTX 3090 GPU. For each 100-frame generation window, audio feature extraction, 4-step Rectified-Flow DiT motion generation, and GAGAvatar rendering take 7.8 ms, 83 ms, and 1.69 s, respectively. The Streaming Audio Scheduler introduces negligible overhead, requiring only 0.029 ms for each mixed-mode scheduling tick. Overall, the pipeline achieves real-time performance and is suitable for interactive deployment.

\section{Ethical Considerations}
\label{sec:supp_ethics} 
EmbodiedHead aims to support positive applications such as conversational agents, virtual education, and assistive communication by providing more natural and engaging head-avatar interactions. The training and evaluation data used in this work are collected from publicly available research datasets and are used only for academic research following their intended research purposes. To protect privacy, we do not collect private videos or personally identifiable information from users, and the user study records only anonymized survey responses.

At the same time, realistic conversational avatar generation may introduce potential misuse risks, such as identity impersonation, misleading synthetic media, or malicious deepfakes. We strictly prohibit the use of this technology for unauthorized identity impersonation, non-consensual content generation, deception, harassment, or any other harmful purposes. When deployed, generated content should be clearly disclosed as synthetic, and watermarking or restricted licenses should be applied to prevent unauthorized or malicious use.

\section{Limitations and Future Work}
\label{sec:supp_limitations}
While EmbodiedHead achieves high-fidelity conversational dynamics, it presents two main limitations. First, the avatar's current listening behaviors are primarily driven by acoustic cues rather than semantic-level dialogue understanding, limiting its intent-driven expressiveness. Second, unlike causal autoregressive (AR) architectures, our diffusion-based framework inherently relies on window-based processing, which introduces a theoretical latency floor. Future work will focus on integrating textual semantics from LLMs to enable context-aware listening responses, as well as exploring hybrid AR-diffusion structures to further minimize streaming latency.

\end{document}